# Advances and Applications of Computer Vision Techniques in Vehicle Trajectory Generation and Surrogate Traffic Safety Indicators

**Mohamed Abdel-Aty, PhD**

Department of Civil, Environmental & Construction Engineering

University of Central Florida, Orlando, FL, 32816, USA

Email: m.aty@ucf.edu

**Zijin Wang***

Department of Civil, Environmental & Construction Engineering

University of Central Florida, Orlando, FL, 32816, USA

Email: zijinwang@knights.ucf.edu

**Ou Zheng, PhD**

Department of Civil, Environmental & Construction Engineering

University of Central Florida, Orlando, FL, 32816, USA

Email: ouzheng1993@knights.ucf.edu

**Amr Abdelraouf, PhD**

Department of Civil, Environmental & Construction Engineering

University of Central Florida, Orlando, FL, 32816, USA

Email: amr.abdelraouf@knights.ucf.edu

## ABSTRACT

The application of Computer Vision (CV) techniques massively stimulates microscopic traffic safety analysis from the perspective of traffic conflicts and near misses, which is usually measured using Surrogate Safety Measures (SSM). However, as video processing and traffic safety modeling are two separate research domains and few research have focused on systematically bridging the gap between them, it is necessary to provide transportation researchers and practitioners with corresponding guidance. With this aim in mind, this paper focuses on reviewing the applications of CV techniques in traffic safety modeling using SSM and suggesting the best way forward. The CV algorithms that are used for vehicle detection and tracking from early approaches to the state-of-the-art models are summarized at a high level. Then, the video pre-processing and post-processing techniques for vehicle trajectory extraction are introduced. A detailed review of SSMs for vehicle trajectory data along with their application on traffic safety analysis is presented. Finally, practical issues in traffic video processing and SSM-based safety analysis are discussed, and the available or potential solutions are provided. This review is expected to assist transportation researchers and engineers with the selection of suitable CV techniques for video processing, and the usage of SSMs for various traffic safety research objectives.

# 1. INTRODUCTION

Traditional traffic safety analysis techniques heavily rely on crash data, which is limited by the data scope considering crashes are rare and it usually takes years to collect sufficient sample to conduct safety analysis and evaluation for a location. With the extensive implementation of roadside closed-circuit television (CCTV) cameras and the advent of unmanned aerial vehicles (UAV), new data sources of video data recording traffic parameters became available. Rich traffic conflict data could be obtained from the vehicle trajectories extracted from videos using computer vision techniques (CV), and thus conflict-based safety analysis have been attracting increasing interest. To conduct conflict-based studies, surrogate safety measures (SSM) are widely adopted due to their capability of measuring temporal or spatial proximity of road users to identify a conflict or a near miss and its severity. Based on SSMs, numerous studies have been conducted for various safety application purposes.

However, the workload from video collection and processing to traffic safety analysis is huge, and many practical issues may be encountered. In most cases, transportation researchers only focused on safety analysis using already available trajectory data or outsourcing the traffic video processing, and their results' validity heavily relied on the trajectory accuracy. Furthermore, relying on existing trajectory data limits the flexibility of conducting safety analysis for customized traffic scenarios (i.e., different time of day and road geometry) and could bring up cost and data privacy issues. In addition, transportation engineers may already have existing infrastructure and systems in place for traffic monitoring, data collection, and analysis. Developing their own CV solutions enables seamless integration with these systems, ensuring compatibility and efficient data processing. For these reasons, traffic safety researchers with video processing capabilities are recently on an increasing demand. Unfortunately, there is not yet literature that systematically introduces solutions for the entire procedure from video processing to safety analysis.

In order to help traffic engineers and researchers to build a concept of applying CV techniques in traffic safety modeling and analysis, a comprehensive review is conducted in this paper with the following objectives:

- Summarize the classic and state-of-the-art CV algorithms for vehicle detection and tracking and present the procedures of applying CV techniques to extract vehicle trajectory from videos.
- Review SSM and their applications on trajectory-based safety research.
- Discuss the practical issues in CV-driven safety research and provide guidance for future research directions.

The schematic diagram of conducting SSM-based safety research using traffic videos is shown in Figure 1, which consists of seven procedures: video collection, video pre-processing, vehicle detection, vehicle tracking, video post-processing, SSM calculation, and safety analysis.

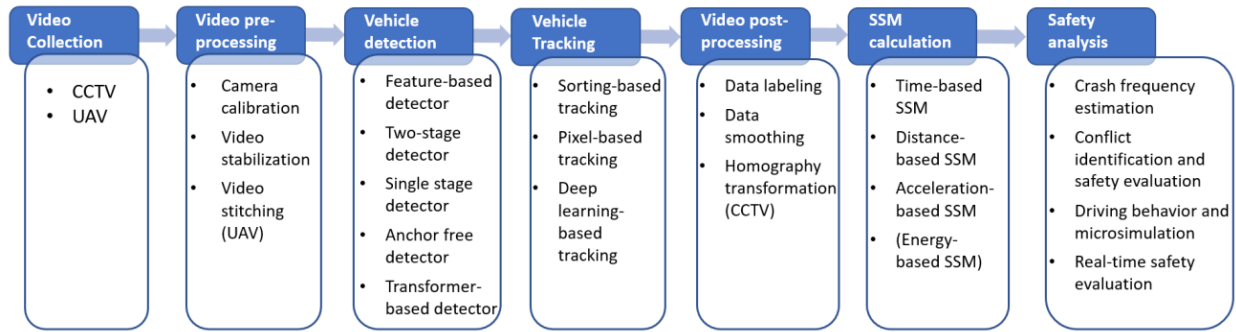

**Figure 1. schematic diagram of video-based traffic safety analysis**

## 2. COMPUTER VISION TECHNIQUES

Typically, to extract vehicle trajectories from a video source, two computer vision techniques are needed: object detection and object tracking. Object detection is the method of identifying, localizing and classifying objects of interest in an image and it is a long-standing challenge in computer vision (Zhao, 2019). In this stage, a video is broken into images by frame, and the vehicles in each image are detected. Major detection models, or detectors, include feature-based, single-stage, two-stage, and anchor-free methods. Object tracking is the method of locating an object across consecutive frames of a video. Each vehicle in the video, as a unique object, is tracked from entering to exiting the video frame. Three common approaches of object tracking are Simple Online and Realtime Tracking (SORT), pixel-based tracking, and deep-learning-based methods.

### 2.1 Object Detection Model

*2.1.1 Feature Based Method*

Featured-based object detection is a method that widely used before the convolutional neural networks (CNNs). It is an object detection method that compute abstractions of image information and judging if each image point belongs to a given type. Optical flow is one of the most popular methods that works in road user classification. It calculates displacement vectors looking for consecutive image pixel matching (Agarwal et al., 2016).

Other conventional feature-based methods include Support Vector Machine (SVM), Histogram of Oriented Gradients (HOG) (Wang et al., 2009), and Scale-Invariant Feature Transform (SIFT) (Lowe, 2004). Overall, the feature-based detection models are classic methods built on relatively simple architecture, and the performance, efficiency, and robustness are not comparable with the more recent deep learning-based models.

*2.1.2 Two-Stage Method*

Since the wide adoption of CNNs and deep learning, numerous deep learning-based detection models were proposed and the performance has been greatly improved. In general, the deep learning-based detectors can be divided into single-stage and two-stage models. The two-stage detector consists of two tasks (1) region proposal and (2) classification and localization. The first task identifies an arbitrary number of objects proposals, and in the second task they are classified and localized.

The first CNN-based two stage detector is Region-based Convolutional Neural Network (R-CNN) that introduced by Girshick et al. (2014). Fast R-CNN (Girshick, 2015), Faster R-CNN (Ren et al., 2015), Mask R-CNN (He et al., 2017), and R-FCN (Dai et al., 2016) are well-known models developed based on R-CNN that improves either accuracy or efficiency.

In general, benefiting from executing two separate tasks for region proposal and localization, better accuracy could be achieved for object localization and recognition. However, it sacrifices the inference speed of the model.

### 2.1.3 Single-Stage Method

Different to the two-stage method, single stage detectors classify and localize an object in a single step using dense sampling. As it skips the task of generating regional proposals, the inference speed of single-stage models is generally higher than two-stage detectors. Hence, it is preferred for real time applications such as vehicle counting.

The famous You Only Look Once (YOLO) algorithm was proposed by Redmon et al. (2016). YOLO is substantially faster than most convolutional neural networks since it conducts classification and bounding box regression in a single step. A later version YOLOv3 provided better performance for smaller objects by utilizing FPN-like feature pyramids (Bochkovskiy et al., 2020). Another single-stage detection model is the Single Shot Multibox Detector (SSD) which uses a single convolutional neural network to detect and classify objects in an image with real-time inference capability (Liu et al., 2016).

### 2.1.4 Anchor Free Method

Most of the popular object detection algorithms of recent times including the abovementioned models use anchor boxes, which are a group of bounding boxes with a specific height and breadth that reflect the size and aspect ratio of the various items that need to be recognized. Anchor free models, which do not use predefined anchor boxes, provide more flexibility in detecting vehicles with different sizes and shapes. Well known anchor free models include FCOS (Tian et al., 2019), FSAF (Zhu et al., 2019), CornerNet (Law & Deng, 2018), ExtremeNet (Zhou et al., 2019), CenterNet (Zhou et al., 2019), and RepPoints (Yang et al., 2019).

Beyond the aforementioned three object detection categories, there exists an additional classification, established upon the recently emerged concept - Transformers. This avant-garde approach was initially proposed within the confines of Natural Language Processing studies, and has lately been gaining recognition and application in object detection. The first transformer-based object detection model is ViT (Dosovitskiy et al., 2020). Other well know Transformers-based detection models are DeTR (Carion et al., 2020), and Swin Transformer (Liu et al., 2021). It worth to mention that the latest Segment Anything Model (SAM) introduced by Meta (Kirillov et al., 2023) is built on the Transformer structure, and it has demonstrated superior performance in object detection, especially in rare object classes (e.g., vehicles with abnormal appearances). Transformer-based models may require significant computational resources and potentially longer training times compared to other detecor categories, depending on the task and implementation.

Figure 2 shows a timeline of the improvement of detection accuracy of the convential detection models on the popular benchmark COCO dataset. It indicates that most recently the variation of CNN-based DINO model performs best. Figure 3 visualizes the inference speed ranking of the detection models. It can be observed that the single-stage models have a higher speed than two-stage R-CNN-based models. Figures 2 and 3 emphasize the tradeoff between accuracy and speed, that will most likely be dependent on the specific objective (could be further addressed with more research and development and higher GPU speed). A summary of the popular models is provided in Table 1. The latest models usually produce better performance in terms of either accuracy or inference speed. Nevertheless, for models proposed around the same time, two-stage detectors have higher detection accuracy than one-stage detectors while sacrificing the detection speed. For real-time vehicle detection or lower accuracy requirement cases (e.g., vehicle counting), Yolo series is the most used model. While for precise vehicle keypoint detection, two-stage models are preferred such as Mask R-CNN and Cascade R-CNN.

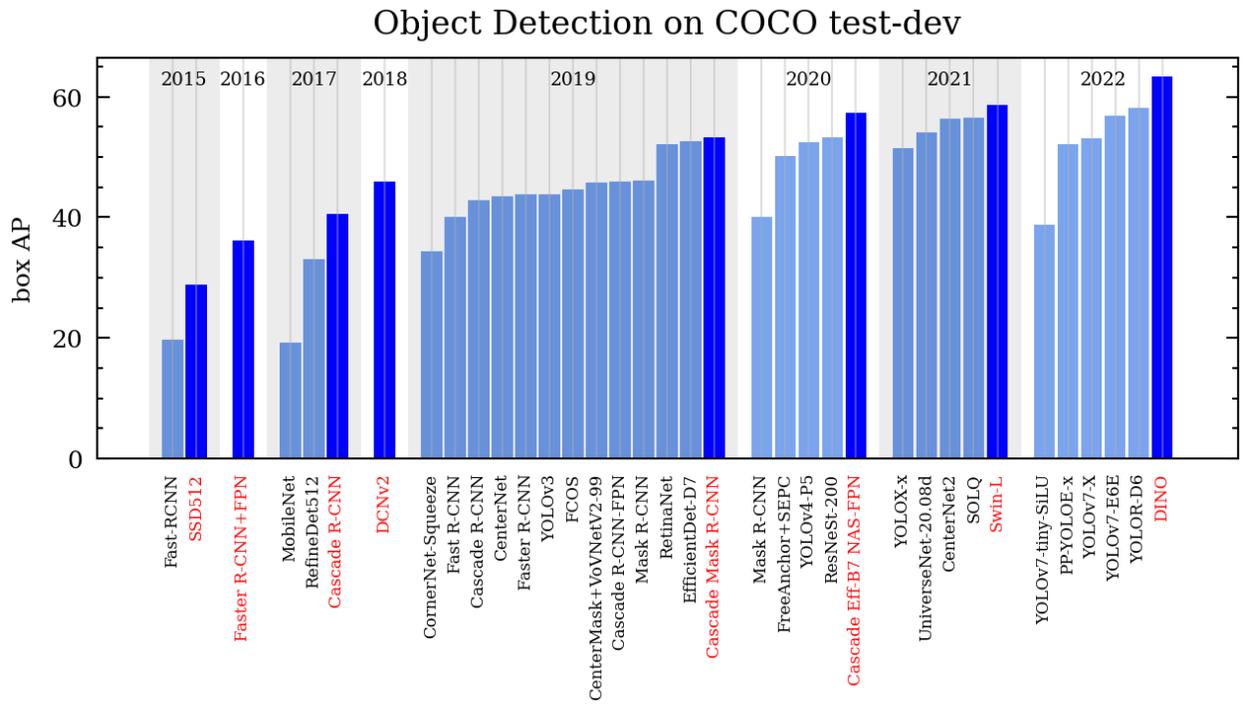

**Figure 2. Accuracy of detection models on COCO test-dev dataset (box AP: bounding box average precision)**

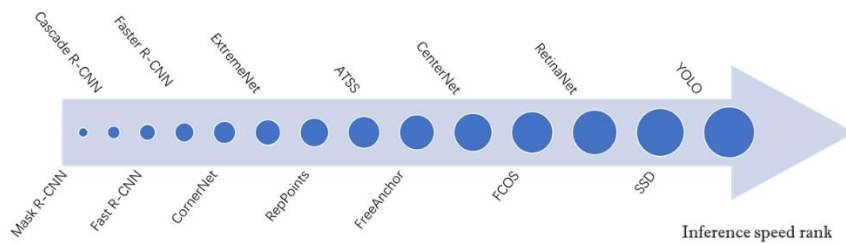

**Figure 3. General inference speed ranking of detection models**

**Table 1. Summary of detection models**

| Algorithm Type | Algorithm | Characteristics |
|---|---|---|
| **One-Stage** (Predict the class label and bounding box coordinates for each object in a single shot, normally with higher inference speed) | SSD (2016) | Multi-scale detection; high memory usage, low interpretability, limited to fixed aspect ratios |
| | YOLO (2016) | Limited accuracy, fixed grid size |
| | YOLOv3 (2018) | Scale-aware for small objects, limited interpretability |
| | YOLOv8 (2023) | Highest speed for real-time detection so far in Yolo series |
| | RetinaNet (2017) | Fast and efficient |
| **Two-Stage** (Detect an object through two stages: the region proposal stage and the classification stage; better localization and recognition accuracy) | Fast R-CNN (2015) | First region-based model, slow |
| | Faster R-CNN (2015) | Faster than Fast R-CNN |
| | Mask R-CNN (2017) | On top of mask R-CNN, capable of instance segmentation tasks |
| | Cascade R-CNN (2020) | Highest accuracy so far |
| **Anchor-free** (Does not use predefined anchor boxes to represent potential object locations in an image; more flexible and adaptable to different object scales and aspect ratios) | CornerNet (2018) | Detecting objects as paired keypoints |
| | ExtremeNet (2020) | Bottom-up object detection by grouping extreme and center points |
| | FCOS (2019) | Predicting a 4D vector, encoding the location of a bounding box |
| | CenterNet (2018) | Identifies visual patterns within individual cropped regions with minimal costs |
| | ATSS (2020) | Automatically select positive and negative samples according to statistical characteristics of object |
| | RepPoints (2019) | Finer representation of objects as a set of sample points useful for both localization and recognition |
| | FreeAnchor (2019) | Break IoU restriction, allowing objects to match anchors in a flexible manner |
| **Feature-based** (Based on the extraction and analysis of local and global features in an image) | Optical flow (1981) | Tracking partially obscured objects, real-time, limited to motion-based detection, less accurate |
| | SVMs | Efficient, robust to noise, requires labeled data, limited to 2D detection |
| | HOG (2005) | Robust to image variations, limited to local features, sensitive to small changes in image |
| | SIFT (1999) | Scale-invariant and rotation-invariant, limited to local features |

## 2.2 Object Tracking Models

*2.2.1 Sorting method*

SORT is a method that tracks an object by matching the detection results across consecutive frames, which could be regarded as a follow-up work of the detection process. For tracking a vehicle, the algorithm first detects and locates a vehicle across frames, and then matches the vehicle based on the Intersection Over Union (IOU) of vehicle bounding boxes in consecutive frames. Conventional sorting methods include DeepSort (Wojke et al., 2017), MOTDT (Chen et al., 2018), QDTrack (Fischer et al., 2022), and ByteTrack (Zhang et al., 2021), etc. However, SORT may fail due to missing detection such as occlusion cases, because the tracking depends on the detection results. This will cause vehicle ID switches under occlusion conditions.

*2.2.2 Pixel-based method*

The pixel-based tracking method tracks an object by directly extracting the movement of each pixel in a video sequence, which means it does not require prior knowledge of detected objects. OpenCV (2023), an open-source computer vision API, has provided researchers and engineers with ready-to-use packages to conduct object tracking. Pixel-based tracking is typically fast and computationally efficient, but it is also more sensitive to the changes in lighting and other factors that can affect the appearance of the pixels in the video. This method is usually used in real-time vehicle tracking applications as well as in tracking vehicle under occlusion conditions (Zheng et al., 2022).

*2.2.3 Deep learning-based tracking*

Another tracking method is deep learning-based tracking, which uses CNN-based model to learn the appearance of an object and track it in the video. Typically, the deep learning model is trained on a large dataset to learn to identify objects and track them by generating a high dimensional feature representation of each frame that is used to estimate the objects' position. Benefitting from the computational power of deep learning, this tracking method can handle complex and dynamic objects with large appearance variations. The limitation of the deep learning-based tracking is that it demands a huge amount of data to obtain high accuracy. However, this issue can be diminished with the advancement in hardware. It needs to mention that the deep learning-based tracking method in this sub-section specifically refers to using deep learning models that directly track an object, rather than tracking based on deep learning-based detection results such as SORT.

## 3. VIDEO PRE-PROCESSING AND POST-PROCESSING

### 3.1 Camera calibration

Detected objects may appear distorted or misplaced due to the uncalibrated camera, and thus it is difficult to estimate their positions and sizes. Radial distortion and tangential distortion are the two main types of distortion. Because of radial distortion, straight lines may seem to be curved. Tangential distortion could result that some portions of the image may appear closer than they should. These issues frequently appeared on CCTV cameras that operate under adverse environmental conditions over extended periods. To undistort a video, camera calibration is needed. The core idea of calibration is to identify multiple sets of 3D real-world points and the corresponding 2D coordinates of these points in the image to calculate the camera matrix and distortion coefficients. A classic two stage calibration method is introduced in Zhang's work (2023), where the external parameters and other parameters are calculated in two separate steps. An automatic calibration approach was proposed by Durai et al. (2022). In their approach, 3D graphics were first transformed to top-down view and then representative points were used to calculate calibration parameters.

### 3.2 Video stabilization

To guarantee vehicle trajectory extraction accuracy, it is important that the video stream is in the best possible state. Video stabilization is a video processing technique for removing undesired camera vibration from a video sequence. The three primary phases of video stabilization algorithms are motion estimation, motion compensation, and finally image composition. Lee et al. proposed a method that uses robust feature trajectory from the video sequence to overcome the vibration issue (Lee et al., 2009). A two-stage solution running in parallel for real-time UAV video stabilization is also introduced (Lim et al., 2019). Pant et al. (Pant et al., 2021) have compared video stabilization techniques that are suitable for UAV data applications.

### 3.3 Video Stitching

For various reasons, the field of view (FOV) of a video may be limited (e.g., fly height restriction for UAV). A method of overcoming the view coverage limitation is to stitch multiple video sources at a same location. In case of transportation, stitching is often a solution to cover longer road segments.

In order to stitch multiple videos, there must be overlapping backgrounds in images from different videos. For instance, when flying two drones to record a road segment, it must make sure that there is an overlapped area between the two video sources. To stitch images, the associations between several overlapping images are first calculated, then the deformed and aligned images are stitched together with the information (Lin et al., 2011). Video stitching is a more complex job than image stitching. By using image stitching algorithms, it typically stitches a few original video frames to create a stitching template that may be used to stitch following frames in accordance with the template. He and Yu introduced a stitching algorithm for UAV video stitching (He & Yu, 2016).

### 3.4 Dataset Labeling

A well-trained model is the cornerstone of the performance of learning-based object model. Although there are many pre-trained vehicle detection models that are open sourced and ready-to-use, the detection accuracy may deteriorate if used for a different scenario (e.g., a new type of vehicle), and cause vehicles to be undetected or misclassified. Thus, it may require human to manually label the incorrectly detected vehicle bounding boxes. To make image data labeling easier and convenient, many labeling tools are available to the researchers, and a review of the labeling tools can be found in (Lima & Reis, 2022).

Data labeling is a tedious process and requires a huge amount of human working hours, and the process can be assisted by active learning techniques, which feedback the manually corrected data into the model training process. Cortés et al. introduced a semi-automatic tracking-based labeling tool to reduce user effort by providing initial baseline labels. Zheng et al. (2022) presented an active learning-based tool for rotating bounding boxes around vehicles.

### 3.5 Coordinate Transformation

For CCTV videos, the roadside view needs to be converted into top-down view to obtain valid vehicle trajectories, which is the homography transformation process. Once the traffic participants are identified and localized in the video frame, the next step is to transform the object location in the image to the local or world coordinate system. Ground plane homography normalization is a common method for translating picture coordinates to world coordinates for transportation monitoring (Zhang, 2000). In another work, Zhang et al. (2019) discussed a mathematical model for transforming the coordinate system. Although this process gives accurate results for UAV data, for roadside cameras the accuracy of detected vehicles can be improved by 3D ground plane estimation. Rangesh and Trivedi introduced a ground plane polling algorithm using 3D object dimension estimation on a 2D image (Rangesh & Trivedi, 2020).

## 4. TRAJECTORY AND SURROGATE-SAFETY-MEASURES

### 4.1 Trajectory Data

The video/image data are collected from the zone of interest of the transportation network, for example an urban intersection, interstate highway or a rural road segment. After running detection and tracking models on the video footage, a sequence of points or bounding boxes that represent the position of vehicles is obtained, which is defined as vehicle trajectory. Two main data sources of vehicle trajectory are UAVs (Menouar et al., 2017) and CCTV or roadside cameras (Eamthanakul et al., 2017).

The adoption of CCTV-based automated detection system appeared as early as 1990, when Autoscope, a wide-area multi-spot video imaging detection system was developed in the United States (Michalopoulos, 1991). Since then, CCTVs have been widely used for traffic state monitoring, vehicle counting, speeding and redlight violation detection, and traffic safety analysis. Most of the recent CCTV cameras provide a wide field of view and capability to rotate to change headings, and support recording continuously if storage permits, making them the prior choice for traffic video collection. On the other hand, CCTV videos may face issues including adverse weather, distortion at the far end, etc. Outputs from the CCTV video are vehicle bounding boxes, which are 2D or 3D rectangles that enclose the entire vehicle and are used to define the vehicle's location and size in the image. Apart from bounding boxes, the detection models may output additional points, denoted as keypoints, that provide more detailed information about the vehicle's shape and features (e.g., headlights, tires). Homography transformation is needed to transform the trajectory from roadside view to top view (Rangesh & Trivedi, 2020; Z. Zhang, 2000).

UAV videos (or drone videos) provide a top view of traffic flow patterns captured by high resolution cameras while positioning the device at a stationary location for a period of time. Since the video is collected from a top view, there is much less distortion in the image and it no longer needs homography transformation. Hence, better accuracy of the trajectory points could be achieved compared to CCTV videos (R. Yuan et al., 2023). The limitations of UAV videos include limited recording time (approximately 20 minutes per flight), and coverage due to flight height restrictions.

With the rapid development of vehicle detection methods and the popularization of UAVs, many research institutes or entities released open datasets of UAV-based vehicle trajectory dataset, that contributed massively to the traffic safety research community. A summary of the available open dataset is shown in Table 2 (updated on 05/06/2023).

**Table 2. Summary of open-source UAV datasets**

| Dataset name | Year | Video FPS | Duration (min) | Road Type |
|---|---|---|---|---|
| NGSIM (Kovvali et al., 2007) | 2005-2006 | 10 | 75 | Highway |
| HighD (Krajewski et al., 2018) | 2017-2018 | 25 | 990 | Highway |
| Interaction (Zhan et al., 2019) | 2019 | 10-30 | 998 | Intersection, expressway |
| CITR & DUT (D. Yang et al., 2019) | 2019 | 29.97 | 18.7 | Intersection |
| Mirror-Traffic | 2019-2020 | 25 | 83.7 | On/off ramp |
| SkyEye (Roy et al., 2020) | 2020 | 50 | 180 | Intersection |
| InD (Bock et al., 2020) | 2020 | 25 | 600 | Intersection |
| RounD (Krajewski et al., 2020) | 2020 | 25 | 360 | Roundabout |
| pNEUMA (Barmpounakis & Geroliminis, 2020) | 2020 | 25 | 3540 | Freeway |
| Ubiquitous Traffic Eye | 2020 | 23.5-25 | 39.75 | Highway |
| High SIM (Shi et al., 2021) | 2021 | 30 | 120 | Highway |
| MAGIC (W. Ma et al., 2022) | 2022 | 25 | 180 | Highway |
| CitySim (O. Zheng et al., 2022) | 2022 | 30 | 1200+ | Highway, intersections, on/off ramps, weaving sections |

### 4.2 Surrogate Safety Measures

Surrogate safety measures (SSM) are the most widely used indicators for traffic safety modeling and analysis, due to their availability and interpretability with the emerging trajectory data (Arun et al., 2021). In this paper, the SSMs that could be derived from video data are summarized and divided into three categories: time-based SSMs, acceleration-based SSMs and distance-based SSMs. There is also another SSM category: energy-based SSMs, which is based on energy loss during an event. However, since most of the energy-based SSMs require the knowledge of vehicle mass and it cannot be obtained from the trajectory, we did not include it in this paper. Also, this paper only focuses on SSMs for vehicle-vehicle conflict, and thus SSMs for vehicle-pedestrian conflict are not discussed, which can be found in (Govinda & Ravishankar, 2022).

*4.2.1 Time-based SSMs*

Time-based SSMs are the most studied and applied SSMs among all the categories. It measures the time proximity for a conflict between two road users. A smaller time gap of an interaction indicates the two road users have a greater chance to come together and form a collision. Time-based SSMs can be further divided into two classes: prediction-based and post-interaction-based.

The first class refers to the SSMs that use the current state to predict future risk of an interaction given certain assumptions, while the second class SSM does not rely on any assumption nor prediction but evaluate the risk or severity of an interaction that has already happened. The most well-known SSM in the first class is Time-to-Collision (TTC), which was first introduced by Hayward in 1971 (Hayward, 1971). The idea of TTC is to measure the time remaining to a potential collision if the two vehicles maintain the current speed and heading. The calculation of TTC is expressed as:

$$TTC_i = \frac{X_{i-1} - X_i - l_i}{v_i - v_{i-1}} \quad (1)$$

Where $i$ and $i$-1 represent the following and leading vehicles, respectively, $X_i$ is the position of vehicle $i$, $l_i$ is the length of vehicle $i$, and $v_i$ is the velocity of vehicle $i$. To identify a conflict from normal interactions, a pre-defined threshold is required, and the TTC value below the threshold is regarded as a conflict. Due to its simplicity and effectiveness in measuring rear-end conflicts, TTC has been widely adopted in the safety evaluation of car following cases and the development of vehicle collision avoidance systems (Z. Wang et al., 2023).

Based on TTC, various time-based SSMs are further developed, including MTTC, TIT, TET, and TA. MTTC improves the precision of TTC in term of estimating the interaction severity by taking the acceleration of vehicle into consideration, which gets rid of the assumption of constant velocity (Ozbay et al., 2008). TIT and TET are well-recognized SSMs that integrate the interaction time duration into the concept of TTC, instead of measuring a conflict severity only using the minimum TTC (Minderhoud & Bovy, 2001). TET is the Time-Exposed-TTC, which represents the duration of an interaction with a TTC value below the threshold. TIT, namely the Time-Integrated-TTC, is the cumulative difference between TTC threshold and current TTC during the conflict period (TTC smaller than the pre-defined threshold) for an interaction. In terms of geometry representation, TIT is the area between the TTC curve and the pre-defined threshold. It reflects the overall risk in terms of conflict duration and severity, and thus provides more detailed information about an interaction. The calculation method of TIT and TET are shown below:

$$TET = \sum_{t=0}^{T} \delta(t) \cdot \tau_{sc}$$
$$\delta(t) = \begin{cases} 0 & else \\ 1 & \forall\, 0 \leq TTC(t) \leq TTC^* \end{cases} \quad (2)$$

$$TIT = \sum_{t=0}^{T} [TTC^* - TTC(t)] \cdot \tau_{sc} \quad (3)$$
$$\forall\, 0 \leq TTC(t) \leq TTC^*$$

where $T$ is the time duration of an interaction, $\tau_{sc}$ is the time scan interval which could be the frame length of the trajectory data, and $TTC^*$ is the TTC threshold value.

TA is another time-based SSM that is derived from the concept of TTC (Perkins & Harris, 1967). Compared to TTC, which consistently calculates the time gap to a potential collision, TA calculates the time gap to a collision from the point when the drivers start to take evasive action. It requires a proper definition of the evasion action and the identification of the event start time, which could vary from different definitions and introduce measurement errors. There are more

SSMs that borrow the concept of TTC, including TTZ (Time-to-Zebra) (Va'rhelyi, 1998), TTL (Time-to-Lane Crossing) (Chin et al., 1992), $TTC_{brake}$ (TTC at brake) (Peng et al., 2017).

Time headway (THW) is another SSM that measures the car following behavior. It is the time gap between the time that the front/rear of the leading vehicle and the front/rear of the following vehicle pass the same point. A small THW represents the following vehicle is following the leading vehicle closely, and may fail to react to sudden brake from the leader. However, it is also pointed out that THW has a weak correlation with crashes, and it cannot distinguish between aggressive driving and conflict situation (Peng, 2018).

The other class of the time-based SSMs is the post-interaction-based SSMs. Post-Encroachment Time (PET) is this class's most widely adopted indicator. PET measures the severity of a conflict from the perspective of post-interaction by calculating the time lapse from the first vehicle vehicles leaves the conflict area to the second vehicle arriving at that area (Allen & Shin, 1978), as shown in Equation 4. Since it is a post-interaction-based SSM, its value can be directly obtained from the trajectory. Same as TTC, a pre-defined threshold is also needed for PET to determine whether an interaction belongs to a conflict or not. It is usually used to measure the risk of angle conflicts or crossing conflicts while it has less validity in the evaluation of car following cases.

$$PET = t_i - t_{i-1} \tag{4}$$

where $t_i$ is the time that the second vehicle arrives at the conflict area, and $t_{i-1}$ is the time that the first vehicle leaves the conflict area.

*4.2.2 Deceleration-based SSMs*

The deceleration-based SSMs measure the risk of an interaction by estimating the deceleration maneuver needed to avoid a potential collision. If a larger deceleration value is required to prevent a crash, then it is considered a more dangerous scenario. The simplest deceleration-based SSM is the deceleration rate measured during an interaction, and a strong relationship between vehicle deceleration and the severity of a conflict at a high speed was found (Malkhamah et al., 2005). Other indicators that are derived directly from the vehicle kinematic features are also used to measure safety including yaw rate, lateral deceleration, and jerk. Deceleration Rate to Avoid the Crash (DRAC) is a famous indicator that is proposed based on the idea of calculating the maximum deceleration rate required to avoid collision under the assumption that the leading vehicle maintains its speed and path (Cooper & Ferguson, 1976). Its mathematical expression is shown below:

$$DRAC = \frac{(v_i - v_{i-1})^2}{2(X_{i-1} - X_i - L_{i-1})} \tag{5}$$

where the parameters here are the same as in the equation for calculating TTC. Based on DRAC, Crash Potential Index (CPI) is developed by considering a vehicle's deceleration ability (José & Cunto, 2008). It is the proportion of the duration that the DRAC exceeds the vehicle's maximum deceleration rate in the interaction period. Deceleration to Safety Time (DST) (Hupfer, 1997) and Unsafe Density (UD) (Barceló et al., 2002) are other deceleration-based SSMs, which measure the risk of a car following case in which the leading vehicle is decelerating.

*4.2.3 Distance-based SSMs*

The distance-based SSMs compare the current gap between leading and following vehicles to the stopping distance of the following vehicles given certain vehicle deceleration assumptions. Rear-end Collision Risk Index (RCRI) are Stopping Distance Index (SDI) are indicators that were proposed for the identification of rear-end conflicts by estimating the minimum distance that the following vehicle requires to prevent colliding with the leading vehicle when the leading vehicle reduces speed with the maximum deceleration rate and stops (Oh et al., 2006). The mathematical expression is shown in Equation 6. RCRI is the SDI over a certain time interval as an exposure.

$$SDI = \begin{cases} 0 \ (safe) & if \ d_L > d_F \\ 1 \ (unsafe) & else \end{cases} \quad (6)$$

where $d_L$ and $d_F$ are the stopping distance for leading and following vehicles, respectively. Time exposed Rear-end Collision Risk Index (TERCRI) is later proposed to measure the risk based on RCRI considering the duration of the interaction, and it is expressed as the accumulation of the timesteps that the RCRI/SDI value equals 1 (Rahman & Abdel-Aty, 2018). Similar to SDI, Potential Index for Collision with Urgent Deceleration (PICUD) also calculates the distance difference between the leading and following vehicles under urgent braking, but SDI produces a binary outcome while PICUD outputs numerical values (Uno et al., 2002). The calculation of PICUD is expressed as:

$$PIUCD = \frac{v_L^2 - v_F^2}{2a} + S_0 - v_F Dt \quad (7)$$

Where $v_L$ and $v_F$ are the velocities of the leading and following vehicles, respectively, $S_0$ is the gap between the two vehicles, $\Delta t$ and $a$ are the pre-defined driver's reaction time and urgent deceleration, respectively. With the same concept as PIUCD, another indicator Difference of Space distance and Stopping distance (DSS) is proposed by adding the friction coefficient into consideration. However, it is less used as most of the collected trajectory data does not contain friction information. Another distance-based SSM is the Proportion of Stopping Distance (PSD), which is the ratio between the current vehicle gap and the stopping distance (Allen & Shin, 1978).

A summary of SSMs is shown in Figure 4.

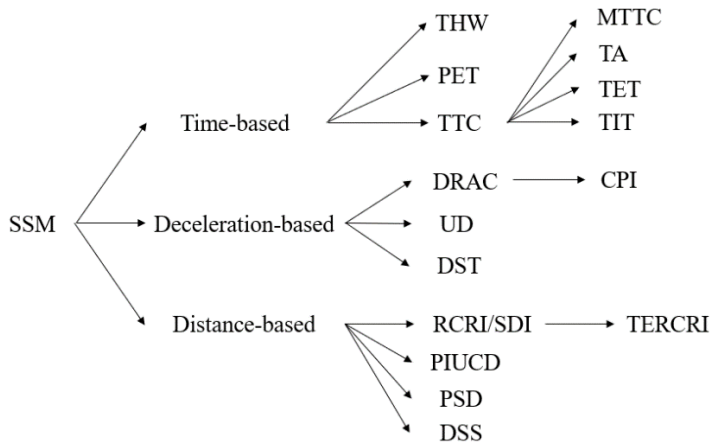

**Figure 4. Summary of SSMs**

## 4.3 SSM-Based Safety Applications

### 4.3.1 Crash frequency estimation

Studies have proven that there are relationships between crashes and conflicts (Tiwari et al., 1998). Hence, the crash frequency in a long horizon could be estimated using the conflict data of a relatively short period. CCTV video data of the traffic at a location with a duration of a few hours to a few days are usually collected, and various SSMs are calculated using the trajectories extracted from the videos, and then expected crash frequency could be estimated based on the distribution of the SSMs.

The most common approach is the Extreme Value Theory (EVT), which directly estimates the crash probability. The idea of EVT approach is building statistical relationship between crashes and conflicts by finding the proportion of extreme values in the SSM observation assuming the SSM follows extreme value distribution. Block Maxima (BM) and Peak Over Threshold (POT) are two models developed based on EVT. The BM model fits a generalized extreme value (GEV) distribution based on the block BM extremes. The observations are aggregated into fixed blocks, and the maxima of each block are treated as extreme, which could be regarded as crashes (Castillo, 2011). Studies using BM approach have been conducted to estimate the crash probability using different SSMs including DRAC, PET, TTC, and MTTC (Fu & Sayed, 2021; Orsini et al., 2019). The POT approach also assumes that the SSM observations follow an extreme value distribution, and use a threshold to determine the proportion of observations that exceed the extreme value, which is regarded as crash-related and thus obtain the crash probability. The crash count over a long period is calculated by the ratio between the estimation horizon and the observation interval multiplied by the crash probability. DRAC, PET, TTC, MTTC, TA, and max deceleration are SSMs commonly used in the literature (Wang et al., 2019; Yang et al., 2021; Zheng & Sayed, 2019).

There is another approach used to estimate crashes by considering the driver heterogeneity and estimating the distribution of driving behavior characteristics, which is defined as uncertainty model (Wang et al., 2021). The general idea of uncertainty models is that human operation errors lead to most of the crashes, and these errors are generated from a small proportion of the driving behavior, which could be modeled based on various SSMs. By adding variances in the driving behavior models, these errors can be statistically quantified to estimate the crash frequency. Microscopic simulation is the common approach applied to obtain the crash count for a future horizon. In the simulation model, driver parameters are calibrated using the estimated distribution based on trajectory data, and the crash count can be generated through repeated simulation (Wang & Stamatiadis, 2016).

### 4.3.2 Conflict identification and safety evaluation

Traffic conflicts are observable events that could end in a crash unless one of the involved parties slows down, changes lanes, or accelerates to a avoid collision. Unlike crashes, which could be strictly identified by the condition that whether the road users collided, conflicts identification normally relies on the use of SSMs. For a set of trajectories, the count and type of conflicts may vary if different SSMs are adopted. Hence, identifying conflicts scientifically and appropriately is crucial for microscopic traffic safety assessment.

Although SSMs are widely used in the estimation and identification of traffic conflicts, its validity in different traffic conditions, road geometries, and conflict types needs to be proven for accurate safety assessment. Many studies have focused on validating the effectiveness of different SSMs on conflict detection using various methods including evasive action identification, trajectory clustering, statistically relating conflicts to crashes. As research have shown the importance of evasive action in identifying traffic conflicts (Tageldin et al., 2017), various studies that use evasive action to assist SSMs for conflicts identification were conducted (Kathuria & Vedagiri, 2020; Lu et al., 2022; Nadimi et al., 2020). The trajectory clustering approach apply clustering techniques to distinguish the conflicts from normal driving through trajectories, and then compare with the SSM-based conflicts to test the validity of various SSMs (Lu et al., 2022; Xie et al., 2019). Also, relating SSM-based conflicts to crashes statistically is another popular approach for SSMs validation. The EVT model, as mentioned above, is commonly used to build the statistical relationship (Orsini et al., 2020).

Another topic is exploring the usage of multiple SSMs for conflict measurement (Wang et al., 2019). Furthermore, extending existing SSMs as new safety indicators are investigated. Aggregated Crash Index (ACI) was proposed based on a tree-search method that could describe eight possible conflict types on freeways (Kuang et al., 2015). Aggregate Conflict Propensity Metric (ACPM) was proposed to better evaluate the crash risk by taking into account the driver's reaction time and braking capacity (Wang & Stamatiadis, 2013). Safety surrogate histogram (SSH) was developed to capture the degree and frequency of traffic conflicts at each intersection approach (Machiani & Abbas, 2016). Aggregated Severe Crash Metric (ASCM) was introduced that could estimate the crash severity (Wang & Stamatiadis, 2014b). Besides, mixture of vehicle kinematic data and SSM could also be an effective method for conflict identification (Xiong et al., 2019).

*4.3.3 Driving behavior and microsimulation*

Growing interests have been observed in using microsimulation to assess road safety by analyzing vehicle trajectories and estimating conflict indicators. However, whether the accuracy of simulated trajectories is precise enough to represent real driving behavior and field-measured conflicts has always been a concern (Essa & Sayed, 2015). Studying driving behavior from the trajectory data and using it to calibrate microsimulation models has been proven effective for better simulation results (Chen et al., 2010; Kesting & Treiber, 2008).

The core of traffic microsimulation is the driving behavior models (e.g., car-following model, lane change model, intersection model), which define the vehicle's behavior in the traffic flow and significantly impact vehicles' interaction. In some safety research that trajectory data are not available, the simulation models were calibrated using mobility measures such as travel time and speed, which dismissed the safety effects of vehicle interactions (Rahman et al., 2019; Wang & Lee, 2021). Although some other research calibrated their simulation model using trajectory data, their calibration objectives are minimizing the error between simulated and field trajectories in terms of time-spatial correlation, and still safety-critical vehicle interactions were not reflected. As stated by Davis et al. (2011), conflicts or crashes are mostly caused by excessive driving behavior, and under-calibrated simulation models may only reflect normal driving behavior and fail to capture and replay excessive behavior. In such case, the driving behavior models calibrated without considering safety matrices may under or over represent safety-critical events and should be used with caution for safety studies.

To evaluate the validity of microsimulation models in conflict simulation, SSMs are considered as an effective tool. The objective of simulation model calibration using SSM is to improve the goodness-of-fit between the simulated and the observed conflicts. A two-step model calibration method is well accepted (Guo et al., 2019). The first step is to reproduce the general traffic environment by considering volume, speed, etc. The second step is to identify driving behavior model parameters that significantly influence conflict simulation by conducting sensitivity analysis. Afterwards, determine the parameter values that minimize the error between simulated and the observed conflicts. Based on this idea, car following model (Guo et al., 2021; Santos, 2015; Sha et al., 2023), lane change model (Bham, 2011; Santos, 2015; Sun & Elefteriadou, 2010), and intersection model (Guo et al., 2019; Mahmud et al., 2019; Sha et al., 2023) were calibrated in the microsimulation models for conflict simulation.

## 5. DISCUSSION

### 5.1 Trajectory Accuracy

Despite the boost in CV models for vehicle trajectory extraction, vehicle detection may encounter several issues because of adverse weather conditions and complicated roadway characteristics. Accordingly, vehicle detection in real-world video sequences is a challenging task and many practical methods have been proposed to address different issues.

Occlusion is a common issue that may cause vehicle detection failure, and it cannot be solved by improving CV techniques as the vehicle is unobservable under occlusion. Birds Eye View (BEV) model, pseudo lidar, and cooperative perception are solutions that are proposed to address this issue. Adverse weather conditions sometimes result in video being blurred and vehicles unobservable, and sensor fusion with radar or lidar could help to overcome such conditions for vehicle detection. Lighting condition is critical for vehicle detection accuracy, as poor lighting can cause issues such as low contrast, shadows, and reflections that can obscure the image and make it difficult for cameras to accurately detect vehicles. Solutions to this include image enhancement methods and sensor fusion techniques. The scale variation issue refers to the changes in the size of objects in an image as they move closer or farther away from the camera. For instance, vehicles at the far end of a CCTV view appear to be smaller. Image pyramids, a technique for resizing an image to multiple different scales, could be applied to address this issue. Rare case, or corner case, is a challenging problem to object detection, especially in autonomous driving. Vehicles with unusual shapes or appearances are typical corner cases and are very likely to be missed by the detection model. For such cases, the direct solution is to label these objects and retrain the detection model. Recent approaches such as unsupervised learning are another solution to the corner cases.

Another concern in vehicle detection is the tradeoff between accuracy and efficiency. Although the advance in CV models focuses on simultaneously improving detection accuracy and inference speed, no model can achieve both the best accuracy and highest speed. For video post processing without time or computational resource concern, CV models with high accuracy and low efficiency are preferred. While for real-time applications, detection accuracy could be relatively sacrificed for better inference speed. The selection of CV models should consider the study or application's purpose and requirement. In general, more recent CV models provide more robust performance in terms of both accuracy and efficiency. Hence, we suggest that researchers closely follow the advances and updates in CV models.

## 5.2 SSM Calculation Method

Although most of the SSMs have clear definitions and mathematical expressions, different calculation methods were adopted according to the characters of trajectory data, which may produce different results and introduce bias. This issue can easily be inferred from the calculation of the most used indicator TTC. First, the TTC calculation method using the vehicle center point omits the effects of vehicle length and obtains a larger gap between leading and following vehicles, and thus tends to underestimate conflicts (Abdelraouf et al., 2022). Many CV algorithms output accurate vehicle bounding boxes, which enable the derivation of true distance between the end of the leader and the front of follower. Hence, using the detected vehicle bounding boxes for TTC calculation could significantly improve the accuracy (Laureshyn et al., 2010). Also, displacement of detected bounding boxes form CCTV videos is a common issue, and pose estimation method was proposed to address this (Abdel-Aty et al., 2022). Second, the concept of TTC is built based on the constant speed and heading assumption. It works well for straight driving cases, however, for car following during turning that vehicle heading is consistently changing, the relative distance between leaders and followers cannot be directly calculated. Various approaches were proposed to address this issue, including predefined grid-based conflict location (Kathuria & Vedagiri, 2020), probabilistic path distribution (St-Aubin et al., 2015; Stipancic et al., 2021), and deep learning-based trajectory prediction (Abdelraouf et al., 2022). Third, if the leading and following vehicles are identified based on relative position with the same lane ID, it may generate extremely small TTC values for the scenario that two vehicles are traveling parallelly on a same lane (such cases are observed in the HighD dataset). Special attention should be paid to avoid using TTC to measure longitudinal safety for such cases.

Furthermore, many other SSMs calculated from trajectory data require certain assumptions which may deteriorate accuracy, such as some deceleration-based (e.g., CPI) or distance-based (e.g., SDI) SSMs need predefined vehicle acceleration capability, and some others (e.g., MTTC) assume the mass of vehicles being the same. Hence, it is suggested to examine the trajectory data carefully prior to SSM calculation from the following perspectives:

(1) Check the definitions and assumptions of the object SSM and examine whether all the assumptions could be fulfilled on the trajectory data.
(2) Determine what features or variables from the trajectory data (e.g., vehicle center point vs. bounding boxes) should be used to calculate SSM following its definitions. If certain features in the trajectory are missing or invalid (e.g., bounding boxes displacement due to video occlusion or distortion), alternative SSMs should be considered.
(3) Examine if there are special cases in the trajectory data that cannot be measured using a specific SSM.

## 5.3 Validity of SSM

Despite the development of many advanced SSMs that have been proven effective in measuring conflicts, more than 65% of the research still adopt TTC and PET due to their simplicity and interpretability (Arun, et al., 2021). However, the power in measuring conflicts may be reduced if the SSM is not suitable for the specific road geometry, conflict type, and traffic condition. Different road geometry leads to various types of conflicts, which have different causes and nature that may need specific SSMs to quantify. For TTC, its capability of reflecting car-following risk has been sufficiently validated, but many studies argue it has limited validity in measuring crossing or angle conflicts. On the contrary, PET is preferred to measure crossing conflicts while it is not recommended for rear-end cases. Other SSMs also have their applicable scenarios and constraints.

Moreover, the traffic conditions could also affect the performance of SSMs (Ding et al., 2023). The rear-end conflicts in high-speed free-flow traffic and low speed or congested traffic should be heterogeneous considering the drivers' gap acceptance at different speeds (Wang, et al., 2022; Wang, et al., 2022). As most SSMs are threshold sensitive, using the same threshold to measure conflicts in different traffic conditions may easily underestimate or overestimate safety. Therefore, additional attention should be paid to the selection of SSM thresholds by taking conflict type and traffic conditions into consideration. Although there is no universal rule for SSM threshold selection, a few methods were proposed and could be adopted. First, to find the threshold to determine conflicts that best statistically relate to crashes. Second, mapping different SSMs into a common and promising conflict indicator. The threshold can be derived by applying mapping functions to the commonly accepted threshold for that conflict indicator. A concern of the above two methods is the selection of the mapping function or distribution. Third, according to field observation of conflicts, determine the conventional thresholds that distinguish conflicts and non-conflicts. Based on the literature reviewed, we summarized the applicable conditions and preferred thresholds for 7 SSMs in Table 3.

**Table 3. Applicable conditions and preferred thresholds for 7 SSM**

| SSM | Threshold | | Pros | Cons |
| --- | --- | --- | --- | --- |
| | Intersection | Freeway | | |
| TTC | 1.5 (1-4) | 3 (1.5-4) | Easy to calculate and interpret; capability of capturing rear-end conflict; most frequently used | Threshold sensitive; rely on constant speed and heading assumption; reduced performance in the congested condition; failure to capture vehicle close-following case |
| PET | 1-3 ($\leq$5) | | Based on observation, no requirement of knowledge of vehicle speed; good interpretability; capable of capturing crossing/angle conflict | Threshold sensitive; only a single value for an interaction, cannot be used to measure risk consistently; little validity of measuring rear-end conflict |
| MTTC | 1-3 ($\leq$4) | 4 ($\leq$5) | Consider the deceleration process of the following vehicle, does not have constant speed assumption; capability of capturing close-following case | Assumes vehicle deceleration and driver reaction time, which may introduce bias |
| THW | | 1 ($\leq$2) | Easy to calculate; reflects the time proximity of the car-following behavior | Only applicable for car-following case; cannot directly identify conflict |
| DRAC | 3.4 ($\leq$4) | 3.4 ($\leq$4) | Good interpretability; easy to calculate; capable of measuring risk for sudden disturbance in traffic | Only applicable for car-following case; hard to determine a universal threshold because vehicle braking performance and road condition varies |

| PSD | 1 | 1 | Good interpretability; relatively easy to measure | The calculation relies on the knowledge of MADR (preferred values include $3.92 m/s^2$, $3.4 m/s^2$, $3.35 m/s^2$) |
| PIUCD | (0)* | (0)* | Good interpretability; measure risk considers two vehicles completely stop; applicable for lane change case | Similar to DRAC and PSD, depend on the selection of deceleration rate; no general adopted threshold |

*: no commonly accepted threshold; the threshold of 0 represents unsafe condition as stated in some literatures

### 5.3 Real-time safety analysis

In the realm of real-time safety evaluation, crash-based analysis has traditionally been the dominant approach (Abdel-Aty et al., 2004, 2023; Hossain et al., 2019; Yuan et al., 2019). However, the recent emergence of drone-based and other vehicle trajectory data has sparked interest in conflict-based real-time safety prediction. However, due to the challenges associated with obtaining accurate trajectory-level data from traffic videos and SSM calculation in real-time, very limited work has been done in this domain.

Real-time vehicle detection and tracking necessitate the utilization of computationally efficient models and algorithms to strike a balance between accuracy and speed. With the rapid advancements in the CV community, advanced algorithms for detection and tracking have been proposed, showcasing improvements in both accuracy and efficiency. Several open-source CV frameworks, such as PaddlePaddle, Open mmlab, Detectron2, PyTorch, TensorFlow, Caffe, and GluonCV, support real-time image or video processing. These frameworks offer downloadable libraries with pre-built functions for detection and tracking, allowing transportation researchers and engineers to easily implement desired algorithms and modify the code to meet specific requirements.

However, real-time conflict identification from vehicle trajectories presents another significant challenge. Most SSM involve calculating values by assessing the temporal and spatial relationships between multiple vehicles. This requires identifying the relative positions of vehicles within the road network, which leads to an exponential increase in processing time with the number of vehicles in the image. Efficient approaches for real-time SSM calculation are still largely unexplored, as existing safety research typically calculates SSM values in post-analysis rather than in real-time.

Despite these challenges, several companies have developed solutions for real-time traffic safety evaluation from videos. Incident detection systems have been developed that aim to identify real-time incidents on the road. Other applications include traffic slowdown detection, pedestrian movement detection and congestion identification, and road usage recording and evaluation. Few provide automated conflict analysis, although the inference speed for real-time applications is yet to be determined.

Researchers are also seeking surrogate methods of real-time safety evaluation through macroscopic traffic state characteristics. This approach aims to investigate the relationship between conflicts and traffic flow features, utilizing the latter to identify unsafe conditions, which can be obtained in real-time traffic detectors. Video-based vehicle trajectories were used to study traffic flow features and conflict distributions to establish their relationships (Yu et al., 2021; Yuan et al., 2022).

### 5.4 Automated Safety Analysis System

With the advances in CV algorithms and safety modeling methodologies, numerous efforts have been made to the safety analysis community using video-based vehicle trajectory data. However, single research only focuses on contributing to a specific problem, while a system level work receives much less attention. There is still a relatively high barrier for transportation researchers and practitioners to start from zero to conduct safety analysis using video data. Hence, frameworks or systems that are capable of automatically conducting safety analysis based on raw video input should be developed to stimulate the research and application of CV techniques in transportation.

The recent adoption of UAV in traffic video collection stimulates the development of video-based automated traffic safety analysis systems. The ARCIS system developed by UCF SST is a representative system that embedded functionalities include video processing, conflict identification, and safety diagnosis based on UAV videos (Wu et al., 2020). The system outputs the rotated bounding rectangles based on the pixel-to-pixel manner masks that are generated by mask R-CNN detection. For conflict identification, PET is calculated for each interaction at the pixel level, and various conflict types can be reported including rear-end, head on, sideswipe, and angle conflict. Furthermore, the system has many other features such as stitching multiple videos, visualization of results among other features.

Traditional workflow from video collection to safety analysis involves several separate tasks, which require step-by-step efforts. The emergence of end-to-end models, the models that learn all the steps between the initial input phase and the final output result, provide new thoughts for developing automated safety analysis systems. In the safety research field, an end-to-end model takes raw videos as input and directly outputs desired safety evaluation metrics while skipping the step of obtaining vehicle trajectories. The introduction of end-to-end models is expected to simplify the conventional video-based safety analysis architecture and improve the efficiency of the safety analysis system for real-time implementation. However, as the model architecture is complex to accomplish the tasks from video process to safety diagnosis in one shot, heavy parameter tuning and training effort are required. Also, enhancing the interpretability of the end-to-end models to provide insights on traffic safety is another challenging task when designing the models. Feature importance selection (Arik & Pfister, 2021), visualization techniques (Gu et al., 2020), sensitivity analysis (Koh & Liang, 2017), and investigating local interpretability (Lundberg & Lee, 2017; Ma et al., 2019) are solutions to improve the interpretability.

## 6. CONCLUSIONS

The application of CV techniques has brought traffic safety analysis into a new era by allowing researchers to study traffic conflicts from a huge amount of vehicle trajectory data. On the other hand, the development of SSMs provides researchers with effective tools to measure traffic conflicts from field data. However, a research gap between the application of CV techniques and traffic safety analysis has been observed. Targeting at this issue, this paper focused on reviewing the application of CV techniques in traffic safety analysis using SSMs. The main contribution of this paper is from five aspects. First, we summarized the CV models that are used for vehicle trajectory extraction from early approaches to state-of-the-art models, including object detection and tracking models. Second, the whole process from video collection to trajectory extraction is explained, and several common video pre-processing and post-processing techniques are presented. Third, a detailed review of SSMs that can be applied to vehicle trajectory data is conducted by dividing the SSMs into three classes: time-based, deceleration-based and distance-based. Fourth, the application of SSMs in conflict measurement and various traffic safety analysis purposes are summarized, including conflict identification, estimation of the relationship between crashes and conflicts, and using trajectory data for microsimulation-based safety research. Last, practical issues in video processing and trajectory-based safety analyses are discussed, and existing solutions and future research directions are suggested. This work is expected to help traffic researchers and engineers build a blueprint of how to use vehicle trajectory data to conduct safety analysis from the perspective of traffic conflicts. Also, the summary and discussion presented in this paper would benefit the researchers in terms of selection of suitable CV techniques and using appropriate SSM(s) for different traffic safety research purposes.